\documentclass[11pt]{article}%
\usepackage{amsmath}
\usepackage{graphicx}%
\usepackage{amsfonts}%
\usepackage{amssymb}
\begin{document}

\title{A Dynamic Programming Algorithm for the Segmentation of Greek Texts}
\author{Fragkou Pavlina}
\maketitle

\abstract{In this paper we introduce a dynamic programming algorithm 
 to perform linear\ text segmentation by global minimization of a segmentation cost
function which consists of: (a)\ within-segment word similarity and (b)\ prior
information about segment length. The evaluation of the segmentation accuracy
of the algorithm on a text collection consisting of Greek texts showed that the algorithm achieves
high segmentation accuracy and appears to be very innovating and promissing.}

\section{Introduction}

\emph{Text segmentation }is an important problem in information retrieval. Its
goal is the division of a text into \emph{homogeneous} (`\emph{lexically
coherent}') segments, i.e. segments exhibiting the following properties: (a)
each segment deals with a particular subject and (b) contiguous segments deal
with different subjects. Those segments can be retrieved from a large database
of unformatted (or loosely formatted) text as being relevant to a query.

This paper presents a dynamic programming algorithm which performs
\emph{linear}\ segmentation \footnote{As opposed to \emph{hierarchical}
segmentation (~\cite{Yaari:97})} by global minimization of a
\emph{segmentation cost}. The \emph{segmentation cost} is defined by a
function consisting of two factors: (a)\ \emph{within-segment word similarity}
and (b)\ \emph{prior information about segment length}. Our algorithm has the
advantage that it can be applied to either large texts - to segment them into
their constituent parts (e.g. to segment an article into sections) - or to a
stream of independent, concatenated texts (e.g. to segment a transcript of
news into separate stories).

For the calculation of the segment \emph{homogeneity }(or alternatively
\emph{heterogeneity}) of a text, several segmentation algorithms using a
variety of criteria have been proposed in the literature. Some of those use
linguistic criteria such as cue phrases, punctuation marks, prosodic features,
reference, syntax and lexical attraction (\cite{{Beeferman:97},{Hirschberg:93}%
,{Passoneau:93}}). Others, following Halliday and Hasan's theory
(\cite{Halliday:76}), utilize statistical similarity measures such as word
cooccurrence. For example the linear discourse segmentation algorithm proposed
by Morris and Hirst (\cite{Morris:91}) is based on \emph{lexical cohesion
relations} determined by use of Roget's thesaurus (\cite{Roget:77}). In the
same direction Kozima's algorithm (\cite{{Kozima:93a},{Kozima:93b}}) computes
the semantic similarity between words using a semantic network constructed
from a subset of the Longman Dictionary of Contemporary English. Local minima
of the similarity scores correspond to the positions of topic boundaries in
the text.

Youmans (\cite{Youmans:91}) and later Hearst (\cite{{Hearst:93},{Hearst:94}})
focused on the similarity between \emph{adjacent} part of texts. They used a
sliding window of text and plotted the number of first-used words in the
window as a function of the window position within the text. In this plot,
segment boundaries correspond to deep valleys followed by sharp upturns. Kan
(\cite{Kan:98}) expanded the same idea by combining word-usage with visual
layout information.

On the other hand, other researchers focused on the similarity between
\emph{all} parts of a text. A graphical representation of this similarity is a
\emph{dotplot}. Reynar (\cite{{Reynar:98},{Reynar:99}}) and Choi
(\cite{{Choi:00},{Choi:01}}) used \emph{dotplots} in conjunction with divisive
clustering (which can be seen as a form of \emph{approximate }and \emph{local
}optimization) to perform \emph{linear} text segmentation. A relevant work has
been proposed by Yaari (\cite{Yaari:97}) who used \emph{divisive /
agglomerative clustering} to perform \emph{hierarchical} segmentation. Another
approach to clustering performs \emph{exact} and \emph{global} optimization by
dynamic programming;\ this was used by Ponte and Croft (\cite{{Ponte:97}%
,{Xu:96}}), Heinonen (\cite{Heinonen:98}) and Utiyama and Isahara
(\cite{Utiyama:01}).

Finally, other researchers use probabilistic approaches to text segmentation
including the use of \emph{Hidden Markov Models} (\cite{Yamron:99},
\cite{Blei:01}). Also Beeferman (\cite{Beeferman:97}) calculated the
probability distribution on segment boundaries by utilizing word usage
statistics, cue words and several other features.

\section{The algorithm}

\label{sec02}

\subsection{Representation}

\label{sec0201}

Suppose that a text contains $T$ sentences and its vocabulary contains $L$
distinct words (e.g words that are not included in the stop list, otherwise
most sentences would be similar to most others). This text can be represented
by a $T\times L$ matrix $F$ defined as follows: for $t=1,2,...,T$ and
$l=1,2,...,L$ we set
\[
F_{t,l}=\left\{
\begin{tabular}
[c]{ll}%
1 & iff l-th word is in t-th sentence\\
0 & else.
\end{tabular}
\ \ \right.
\]

The \emph{sentence similarity matrix} of the text is a $T\times T$ matrix $D$
where for $s,t=1,2,...,T$ we set
\[
D_{s,t}=\left\{
\begin{tabular}
[c]{lll}%
1 & if & $\sum_{l=1}^{L}F_{s,l}F_{t,l}>0;$\\
0 & if & $\sum_{l=1}^{L}F_{s,l}F_{t,l}=0.$%
\end{tabular}
\ \ \right.
\]
This means that $D_{s,t}$ = 1 if the $s$-th and $t$-th sentence have at least
one word in common. Every part of the original text corresponds to a submatrix
of $D$. It is expected that submatrices which correspond to actual segments
will have many sentences with words in common, thus will contain many `ones'.
In Figure 1 we give a \emph{dotplot} of a matrix corrpesponding to a
91-sentences text. Ones are plotted as black squares and zeros as white
squares. Further justification for the use of this similarity matrix and
graphical representation can be found in \cite{Petridis:01},
\cite{Kehagias:02}, \cite{{Reynar:98},{Reynar:99}}, \cite{Choi:00} and
\cite{Choi:01}.

We make the assumption that segment boundaries always occur at the end of
sentences. A segmentation\emph{\ }of a text is a partition of the set
$\{1,2,...,T\}$ into $K$ subsets (i.e. \emph{segments, }where $K$ is a
variable number) of the form $\{1,2,...,t_{1}\}$, $\{t_{1}+1,t_{1}%
+2,...,t_{2}\}$, ..., $\{t_{K-1}+1,t_{K-1}+2,...,T\}$ and can be represented
by a vector $\ \mathbf{t}=(t_{0},t_{1},...,t_{K})$, where $t_{0}%
,t_{1},...,t_{K}$\emph{\ }are the \emph{segment boundaries }corresponding to
the last sentence of each subset.

\subsection{Dynamic Programming}

\label{sec0202}

Dynamic programming as a method guarantees the optimality of the result with
respect to the input and the parameters. Following the approach of Heinonen
(\cite{Heinonen:98}) we use a dynamic programming algorithm which decides the
locations of the segment boundaries by calculating the globally optimal
splitting $\mathbf{t}$ on the basis of a similarity matrix (or a curve), a
preferred fragment length and a cost function defined. Given a similarity
matrix $D$ and the parameters $\mu$\textit{, }$\sigma$, $r$ and $\gamma$ (the
role of each of which will be described in the sequel) the dynamic programming
algorithm\ tries to minimize a \emph{segmentation cost function }%
$J(\mathbf{t};\mu,\sigma,r,\gamma)$ with respect to $\mathbf{t}$ (here
$\mathbf{t}$ is the independent variable which is actually a vector specifying
the boundary position of each segment and the number of segments $K$ while
$\mu,\sigma,r, \gamma$ are parameters) which is defined in equation (1).%
\begin{equation}%
\begin{tabular}
[c]{l}%
$J(\mathbf{t};\mu,\sigma,r,\gamma)=\sum_{k=1}^{K}\gamma{\cdot}\frac{\left(
t_{k}-t_{k-1}-\mu\right)  ^{2}}{2\cdot\sigma^{2}}-(1-\gamma){\cdot}%
\frac{\sum_{s=t_{k-1}+1}^{t_{k}}\sum_{t=t_{k-1}+1}^{t_{k}}D_{s,t}}{\left(
t_{k}-t_{k-1}\right)  ^{r}}$%
\end{tabular}
\label{equ02cont}%
\end{equation}

Hence the sum of the costs of the $K$ segments constitutes the total
segmentation cost; the cost of each segment is the sum of the following two
terms (with their relative importance weighted by the parameter $\gamma$): \smallskip

\begin{enumerate}
\item The term \emph{\ }$\frac{\left(  t_{k}-t_{k-1}-\mu\right)  ^{2}}%
{2\cdot\sigma^{2}}$ corresponds to the length information measured as the
deviation from the average segment length. In this sense, $\mu$ and $\sigma$
can be considered as the mean and standard deviation of segment
length\ measured either on the basis of words or on the basis of sentences
appearing in the text's segments and can be estimated from \emph{training
data}.

\item The term $\frac{\sum_{s=t_{k-1}+1}^{t_{k}}\sum_{t=t_{k-1}+1}^{t_{k}%
}D_{s,t}}{\left(  t_{k}-t_{k-1}\right)  ^{r}}$ corresponds to (word)
similarity between sentences. The numerator of this term is the total number
of ones in the $D$ submatrix corresponding to the $k $-th segment. In the case
where the parameter $r$ is equal to 2, $\left(  t_{k}-t_{k-1}\right)  ^{r}$
correspond to the area of submatrix and the above fraction corresponds to
`segment density'. A `generalized density' is obtained when $r\neq2$ and
enables us to control the degree of influence of the surface with regard to
the `information' (i.e. the number of ones) included in it. Strong
intra-segment similarity (as measured by the number of words which are common
between sentences belonging to the segment) is indicated by large values of
$\frac{\sum_{s=t_{k-1}+1}^{t_{k}}\sum_{t=t_{k-1}+1}^{t_{k}}D_{s,t}}{\left(
t_{k}-t_{k-1}\right)  ^{r}}$ , irrespective of the exact value of $r$.
\end{enumerate}

\smallskip

Segments with high density and small deviation from average segment length
(i.e. a small value of the corresponding $J(\mathbf{t};\mu,\sigma,r,\gamma) $
\footnote{Small in the \emph{algebraic }sense; $J(\mathbf{t};\mu
,\sigma,r,\gamma)$ can take both positive and negative values.}) provide a
`good' segmentation vector $\mathbf{t}$. The \emph{global} minimum of
$J(\mathbf{t};\mu,\sigma,r,\gamma)$ provides the \emph{optimal }segmentation
$\widehat{\mathbf{t}}$. It is worth mentioning that the optimal $\widehat
{\mathbf{t}}$ specifies both the optimal number of segments $K$ and the
optimal positions of the segment boundaries \emph{\ }$t_{0},t_{1},...,t_{K.}$
In the sequel, our algorithm is presented in a form of pseudocode.

\bigskip

\textbf{Dynamic Programming Algorithm}

\textbf{------------------------------------------------------}

\textbf{Input: }The $T\times T$ similarity matrix $D$; the parameters $\mu$,
$\sigma$, $r$, $\gamma.$

\textbf{Initilization}

For $t=1,2,..T$

\qquad$Sum=0$

\qquad For $s=1,2,...,t-1$

\qquad\qquad$Sum=Sum+$ $D_{s,t}$

\qquad End

\qquad$S_{s,t}$ $=$ $\frac{Sum}{(t-s)^{r}}$

End

\textbf{Minimization}

$C_{0}$ $=0,$ $Z_{0}=0$

For $t=1,2,,T$

\qquad$C_{t}$ = $\infty$

\qquad For $s=1,2,...,t-1$

\qquad\qquad If $C_{s}+S_{s,t}+$ $\frac{(t-s-\mu)^{2}}{2\sigma^{2}}$ $\leq
C_{t}$

\qquad\qquad\qquad\ $C_{t}=C_{s}+S_{s,t}+\frac{(t-s-\mu)^{2}}{2\sigma^{2}}$

\qquad\qquad$\ \qquad Z_{t}=s$

\qquad\qquad EndIf

\qquad End

End

\textbf{BackTracking}

$K=0$ $,s_{k}=T$

While $Z_{s_{k}}>0$

\qquad\qquad$k=k+1$

\qquad\qquad$s_{k}=Z_{s_{k-1}}$

End

$K=K+1,$ $Z_{k}=0,$ $\widehat{t}_{0}=0$

For $k=1,2,...,K$

$\qquad\widehat{t}_{k}=s_{K-k}$

End

\textbf{Output: }The optimal segmentation vector $\widehat{\mathbf{t}%
}=(\widehat{t}_{0},\widehat{t}_{1},...,\widehat{t}_{K}).$

\begin{center}
\textbf{------------------------------------------------------}
\end{center}

\section{Evaluation}

\label{sec03}

\subsection{Measures of Segmentation Accuracy}

\label{sec0301}

The performance of our algorithm was evaluated by three
indices:\ \emph{Precision, Recall }and Beeferman's $P_{k}$ \emph{metric}.
\emph{Precision} and \emph{Recall} measure segmentation \emph{accuracy}. For
the segmentation task, \emph{Precision} is defined as `the number of the
estimated segment boundaries which are actual segment boundaries' divided by
`the number of the estimated segment boundaries'. On the other hand,
\emph{Recall }is defined as `the number of the estimated segment boundaries
which are actual segment boundaries' divided by `the number of the true
segment boundaries'. High segmentation accuracy is indicated by high values of
\emph{both Precision} and \emph{Recall}. However, these two indices have some
shortcomings. First, high \emph{Precision} can be obtained at the expense of
low \emph{Recall} and conversely. Additionally, those two indices penalize
equally every inaccurately estimated segment boundary whether it is near or
far from a true segment boundary.

An alternative measure $P_{k}$ which overcomes the shortcomings of
\emph{Precision} and \emph{Recall} and measures segmentation \emph{in}accuracy
was introduced recently by Beeferman et al. (\cite{Beeferman:97}).
Intuitively, $P_{k}$ measures the proportion of `sentences which are wrongly
predicted to belong to the same segment (while actually they belong in
different segments)' or\textit{\ }`sentences which are wrongly predicted to
belong to different segments (while actually they belong to the same
segment)'. $P_{k}$ is a measure of how well the true and hypothetical
segmentations agree (with a low value of $P_{k}$ indicating high accuracy
(\cite{Beeferman:97})). $P_{k}$ penalizes near-boundary errors less than
far-boundary errors. Hence $P_{k}$ evaluates segmentation accuracy more
accurately than \emph{Precision} and \emph{Recall}.

\subsection{Experiments}

For the experiments, we use a text collection compiled from a corpus
comprising of text downloaded from the
website\texttt{\ http://tovima.dolnet.gr} of the newspaper entitled `To Vima'.
This newspaper contains articles belonging to the following categories: 1)
Editorial, diaries, reportage, politics, international affairs, sport reviews
2) cultural supplement 3) Review magazine 4) Business, finance 5) Personal
Finance 6) Issue of the week 7) Book review supplement 8) Art review
supplement 9) Travel supplement. Stamatatos et al. (~\cite{Stamatatos:01})
constructed a corpus collecting texts from supplement 2) which includes essays
on science, culture, history etc. They selected 10 authors from the above set
without taking any special criteria into account. Then 30 texts of each author
were downloaded from the website of the newspaper as shown in the table
below:
%\newline 

\begin{center}%
\begin{tabular}
[c]{|r|r|}\hline
\textbf{Author } & \textbf{Thematic Area }\\\hline
Alachiotis & Biology\\\hline
Babiniotis & Linguistics\\\hline
Dertilis & History,Society\\\hline
Kiosse & Archeology\\\hline
Liakos & History,Society\\\hline
Maronitis & Culture,Society\\\hline
Ploritis & Culture,History\\\hline
Tassios & Technology,Society\\\hline
Tsukalas & International Affairs\\\hline
Vokos & Philosophy\\\hline
\end{tabular}

\textbf{Table 1}:

List of Authors and Thematic Areas dealt by each of those.
\end{center}

No manual text preprocessing nor text sampling was performed aside from
removing unnecessary heading irrelevant to the text itself. All the downloaded
texts were taken from the issues published from 1997 till early 1999 in order
to minimize the potential change of the personal style of an author over time.
Further details can be found in \cite{Stamatatos:01}.

The preprocessing of the above texts was made using the morphosyntactic tagger
(better known as part-of-speech tagger) developed by Giorgos Orphanos
(\cite{{Orphanos:99a},{Orphanos:99c}}). The aforementioned tagger is a POS
tagger for modern Greek (a high inflectional language) which is based on a
Lexicon capable of assigning full morphosyntactic attributes (i.e. POS,
Number, Gender, Tense, Voice, Mood and Lemma) to 876.000 Greek word forms.
This Lexicon was used to build a tagged corpus capable of showing off the
behavior of all POS ambiguity schemes present in the Modern Greek (e.g.
Pronoun-Clitic-Article, Pronoun-Clitic, Adjective-Adverb, Verb-Noun, etc) as
well as the characteristics of unknown words. This corpus was used for the
induction of decision trees, which along with the Lexicon are integrated into
a robust POS tagger for Modern Greek texts.

The tagger architecture consists of three parts: the Tokenizer, the Lexicon
and finally the Disambiguator and Guesser. Raw text passes through the
Tokenizer, where it is converted to a stream of tokens. Non-word tokens (e.g.
punctuation marks, numbers, dates etc.) are resolved by the Tokenizer and
receive a tag corresponding to their category. Word tokens are looked up in
the Lexicon and those found receive one or more tags. Words with more than one
tags and those not found in the Lexicon pass through the
Disambiguator/Guesser, where the contextually appropriate tag is decided.

The Disambiguator/Guesser is a `forest' of decision trees, one tree for each
ambiguity scheme present in Modern Greek and one tree for unknown guessing.
When a word with two or more tags appears, its ambiguity scheme is identified.
Then, the corresponding decision tree, is selected, which is traversed
according to the values of the morphosyntactic features extracted from
contextual tags. This traversal returns the contextually appropriate POS along
with its corresponding lemma. The ambiguity is resolved by eliminating the
tag(s) with different POS than the one returned by the decision tree. The POS
of an unknown word is guessed by traversing the decision tree for unknown
words, which examines contextual features along with the word ending and
capitalization and returns an open class POS and the corresponding lemma.

\subsubsection{Preprocessing}

For the experiments we use the texts taken from the collection compiled from
the corpus of the newspaper `To Vima'. Each of the 300 texts of the collection
of articles compiled from this newspaper is preprocessed using the POS tagger
created by G. Orphanos. More specifically, every word in the text was
substituted by its lemma, determined by the tagger. Punctuation marks, numbers
and all words were removed except words that are either nouns, verbs,
adjectives or adverbs. For those words that their lemma was not determined by
the tagger, due to the fact that those words were not contained in the Lexicon
used for the creation of the tagger, no substitution was made and the words
were used as they were. The only information that was kept was the end of each
sentence appearing in each text. We next present two suites of experiments.
The difference between those suites lies in the length of segments created and
the number of authors used for the creation of the texts to segment, where
each text being a concatenation of ten text segments.

\subsubsection{1st suite of experiments}

In the first suite of experiments, our collection consists of 6 datasets:
Set0, ..., Set5. The difference between those datasets lies in the number of
authors used for the generation of the texts to segment and consequently the
number of texts used from the collection. The table below contains the
aforementioned information.
%\newline 

\begin{center}%
\begin{tabular}
[c]{|r|r|r|}\hline
{\small Set} & {\small Authors } & {\small No. of texts per set}\\\hline
{\small 0} & {\small Kiosse, Alachiotis} & {\small 60}\\\hline
{\small 1} & {\small Kiosse, Maronitis} & {\small 60}\\\hline
{\small 2} & {\small Kiosse, Alachiotis, Maronitis} & {\small 90}\\\hline
{\small 3} & {\small Kiosse, Alachiotis, Maronitis, Ploritis} & {\small 120}%
\\\hline
{\small 4} & {\small Kiosse, Alachiotis, Maronitis, Ploritis, Vokos } &
{\small 150}\\\hline
{\small 5} & {\small All Authors} & {\small 300}\\\hline
\end{tabular}
\qquad\qquad

\textbf{Table 2}{\small :}

{\small List of the sets complied in the 1st suite of experiments and the
author's texts used for each of those.}
\end{center}

For each of the above sets, we constructed four subsets. The difference
between those subsets lies in the range of the sentences appearing in each
segment for every text. If a and b correspond to the lower and higher values
of sentences consisting each segment, we have used four different pairs:
(3,11), (3,5), (6,8) and (9,11). In every dataset, before generating any of
the texts to segment, each of which containing 10 segments, we selected the
authors, whose texts will be used for this generation. If X is the number of
authors contributing to the generation of the dataset, for all datasets, each
text is generating according to the following procedure (which guarantees that
each text contains ten segments): \smallskip

\begin{enumerate}
\item For j=1,2,..., 10 where j corresponds to the j-th out of the 10 segments
of the generated text.

\item For I=1,...,X a random integer -corresponding to an author - is generated.

\item For k=1,2,...,30 a random integer corresponding to the texts belonging
to the selected author I is generated.

\item For l=a,...,b, a random integer corresponding to the number of
consecutive sentences extracted from text k (starting at the first sentence of
the text) is generated.
\end{enumerate}

\smallskip

For every subset, using the procedure described above, we generated 50 texts.
As it was mentioned before, our algorithm uses four parameters $\mu,\sigma,
\gamma$ and $r,$ where $\mu$ and $\sigma$ can be interpreted as the average
and standard deviation of segment length; it is not immediately obvious how to
calculate the optimal values of each of parameters. A procedure for
determining appropriate values of $\mu,\sigma,\gamma$ and $r$ was introduced
using \emph{training data} and a \emph{parameter validation } procedure. Then
our algorithm is evaluated on (previously unseen)\ \emph{test data}. More
specifically, for each of the datasets Set0,..., Set5 and each of their
subsets we perform the procedure described in the sequel: \smallskip

\begin{enumerate}
\item Half of the texts in the dataset are chosen randomly to be used as
training texts; the rest of the samples are set aside to be used as test texts.

\item Appropriate $\mu$ and $\sigma$ values are determined using all the
training texts and the standard statistical estimators.

\item Parameter $\gamma$ is set to take the values 0.00, 0.01, 0.02, ... ,
0.09, 0.1, 0.2, 0.3, ... , 1.0 and $r$ to take the values 0.33, 0.5, 0.66,1.
This yields 20$\times$4=80 possible combinations of $\gamma$ and $r$ values.
Appropriate $\gamma$ and $r$ values are determined by running the segmentation
algorithm on all the training texts with the 80 possible combinations of
$\gamma$ and $r$ values; the one that yields the minimum $P_{k}$ value is
considered to be the optimal ($\gamma$, $r$) combination.

\item The algorithm is applied to\ the test texts using previously estimated
$\gamma,$ $r,$ $\mu$ and $\sigma$ values.
\end{enumerate}

\smallskip

An idea of the influence of $\gamma$ and $r$ on $P_{k}$ of the first suite of
experiments can be observed in Figures 2-5 (corresponding to subsets '3-11',
'3-5', '6-8' and '9-11' of Set5). In those figures \emph{Exp 1 }refers to the
first suite of experiments.

The above procedure is repeated five times for each of the six datasets and
the resulting values of \emph{Precision, Recall} and $P_{k}$ are averaged. The
performance of our algorithm (as obtained by the validated parameter values)
is presented in Table 3.

\subsubsection{2nd Suite of experiments}

In the second suite of experiments we used the same collection of texts
compiled from the corpus of the newspaper `To Vima'. The difference between
those two suites lies in the way of generating the texts used for training and
for testing. In this suite of experiments, we used all the available (300)
texts of the collection of the Greek corpus, which means all the available
authors. We constructed a single dataset containing 200 texts. Half of them
were used for training while the rest of them were used for testing. Each of
the aforementioned texts was generated according to the following procedure
(which guarantees that each text contains 10 segments):

\begin{enumerate}
\item For j=1,2,..., 10 where j corresponds to the j-th out of the 10 segments
of the generated text.

\item For I=1,...,X a random integer -corresponding to an author - is generated.

\item For k=1,2,...,30 a random integer corresponding to the texts belonging
to the selected author I is generated. The selected text is read and scanned
in order to determine the number of paragraphs that consists it. If Z is the
number of paragraphs that consists it then:

\item For l = 1,2,...,Z a random integer- corresponding to the number of
paragraphs appearing in text k - is generated.

\item For m = 1,..., Z-l, a random integer - corresponding to the "starting
paragraph"- is generated. Thus the segment contains all the paragraphs of text
k starting from paragraph m and ending at the paragraph m+l.
\end{enumerate}

\begin{center}%
\[%
\begin{tabular}
[c]{|l|c|c|c|c|}\hline
{\small \emph{1st suite of Experiments}} & {\small (3,11)} & {\small (3,5)} &
{\small (6,8)} & {\small (9,11)}\\\hline
{\small Set0 Precision} & \multicolumn{1}{|r|}{{\small 70,65\%}} &
\multicolumn{1}{|r|}{{\small 86,82\%}} & \multicolumn{1}{|r|}{{\small 96,44\%}%
} & \multicolumn{1}{|r|}{{\small 93,33\%}}\\\hline
{\small Set0 Recall } & \multicolumn{1}{|r|}{{\small 71,11\%}} &
\multicolumn{1}{|r|}{{\small 87,11\%}} & \multicolumn{1}{|r|}{{\small 96,44\%}%
} & \multicolumn{1}{|r|}{{\small 93,33\%}}\\\hline
{\small Set0 Beeferman} & \multicolumn{1}{|r|}{{\small 14,04\%}} &
\multicolumn{1}{|r|}{{\small 6,20\%}} & \multicolumn{1}{|r|}{{\small 0,82\%}}
& \multicolumn{1}{|r|}{{\small 0,84\%}}\\\hline
{\small Set1 Precision} & \multicolumn{1}{|r|}{{\small 63,86\%}} &
\multicolumn{1}{|r|}{{\small 82,98\%}} & \multicolumn{1}{|r|}{{\small 91,11\%}%
} & \multicolumn{1}{|r|}{{\small 94,67\%}}\\\hline
{\small Set1 Recall } & \multicolumn{1}{|r|}{{\small 67,11\%}} &
\multicolumn{1}{|r|}{{\small 83,56\%}} & \multicolumn{1}{|r|}{{\small 91,11\%}%
} & \multicolumn{1}{|r|}{{\small 94,67\%}}\\\hline
{\small Set1 Beeferman} & {\small 15,82\%} & {\small 8,47\%} & {\small 2,81\%}
& {\small 0,98\%}\\\hline
{\small Set2 Precision} & {\small 71,14\%} & {\small 90\%} & {\small 91,11\%}
& {\small 92,44\%}\\\hline
{\small Set2 Recall } & {\small 60,89\%} & {\small 89,78\%} & {\small 91,11\%}
& {\small 92,44\%}\\\hline
{\small Set2 Beeferman} & {\small 14,42\%} & {\small 3,45\%} & {\small 2,15\%}
& {\small 1,247\%}\\\hline
{\small Set3 Precision} & {\small 59,99\%} & {\small 84,44\%} &
{\small 86,22\%} & {\small 91,11\%}\\\hline
{\small Set3 Recall } & {\small 58,67\%} & {\small 83,56\%} & {\small 86,22\%}
& {\small 91,11\%}\\\hline
{\small Set3 Beeferman} & {\small 17,93\%} & {\small 7,36\%} & {\small 3,28\%}
& {\small 1,45\%}\\\hline
{\small Set4 Precision} & {\small 57,99\%} & {\small 85\%} & {\small 88,89\%}
& {\small 91,11\%}\\\hline
{\small Set4 Recall } & {\small 51,11\%} & {\small 84,89\%} & {\small 88,89\%}
& {\small 91,11\%}\\\hline
{\small Set4 Beeferman} & {\small 17,38\%} & {\small 6,76\%} & {\small 2,65\%}
& {\small 1,39\%}\\\hline
{\small Set5 Precision} & {\small 65,74\%} & {\small 81,56\%} &
{\small 89,33\%} & {\small 88,89\%}\\\hline
{\small Set5 Recall } & {\small 61,78\%} & {\small 81,78\%} & {\small 89,33\%}
& {\small 88,89\%}\\\hline
{\small Set5 Beeferman} & {\small 14,54\%} & {\small 6,49\%} & {\small 3,57\%}
& {\small 1,86\%}\\\hline
\end{tabular}
\]

\textbf{Table 3}

{\small Exp.Suite 1: The Precision, Recall and Beeferman's metric values for
the datasets Set0, Set1, Set2, Set3, Set4 and Set5, using sentences as a unit
of segment, obtained by a validation procedure. }
\end{center}

From the aforementioned method of generating texts, it is obvious that, the
200 generated texts for segmentation are larger - in length - than those
generated during the first suite of experiments. Thus the segmentation of such
texts consists a more difficult problem. We used the same validation procedure
as before with the same values for the parameters r and $\gamma$. The obtained
validated results are listed in the table below:
%\newline 

\begin{center}%
\begin{tabular}
[c]{|l|c|}\hline
{\small \emph{2nd suite of Experiments}} & \\\hline
{\small Precision} & \multicolumn{1}{|r|}{{\small 60,60\%}}\\\hline
{\small Recall } & \multicolumn{1}{|r|}{{\small 57,00\%}}\\\hline
{\small Beeferman} & \multicolumn{1}{|r|}{{\small 11,07\%}}\\\hline
\end{tabular}

\textbf{Table 4}

{\small Exp.Suite 2: The Precision, Recall and Beeferman's metric values for
the unique dataset using paragraphs as a unit of segment obtained by a
validation procedure. }
\end{center}

\section{Discussion}

Our algorithm was previously tested on Choi's data collection
(\cite{Kehagias:03}), which contains english texts, achieving significantly
better results than the ones previously reported in \cite{{Choi:00},
{Choi:01}} and \cite{Utiyama:01}. Since the collection used here has not been
previously used in the literature for the purpose of text segmentation, we
cannot provide a direct comparative assessment. However, the performance
obtained is comparable and in most cases better than the corresponding on the
Choi's collection, even though, for several cases the problem dealt by our
algorithm is more difficult. The difficulty lies in the fact that, the
thematic area dealt by several authors is very similar (see Table 1). One of
the reasons for the high segmentation accuracy is the robustness of the POS
tagger used. We have observed that, in general, the tagger fails to find the
tag and lemma of very technical words. The use of them as they appear in the
original text, does not have a negative impact on the segmentation accuracy.
The robustness of our algorithm is also indicated by the performance obtained
at the second suite of experiments where the segment length is bigger and the
deviation from the average length is high. Even in that case our algorithm
achieved very high results. This is the result of the combination of the
following facts: First, the use of the segment length term in the cost
function seems to improve segmentation accuracy significantly. Second, the use
of `generalized density' ($r\neq2$) appears to significantly improve
performance. Even though the use of `true density' ($r=2$) appears more
natural, the best segmentation performance (minimum value of $P_{k}$) is
achieved for significantly smaller values of $r$. This performance in most
cases is improved when using appropriate values of $\mu,\sigma,\gamma$ and $r$
derived from training data and parameter validation.

Finally, it is worth mentioning that our approach is `global' in two respects.
First, sentence similarity is computed globally through the use of the $D$
matrix and \emph{dotplot}. Second, this global similarity information is also
optimized globally by the use of the dynamic programming algorithm. This is in
contrast with the local optimization of global information (used by Choi) and
global optimization of local information (used by Heinonen).

It is worth mentioning that, the computational complexity of our algorithm is
comparable to that of the other methods (namely $O(T^{2})$ where $T$ is the
number of sentences). Finally, our algorithm has the advantage of
automatically determining the optimal number of segments.

\section{Conclusion}

We have presented a dynamic programming algorithm which performs text
segmentation by global minimization of a segmentation cost consisting of two
terms:\ within-segment word similarity and prior information about segment
length. The performance of our algorithm is quite satisfactory considering
that it yields a high performance in a text collection containing Greek texts.
In the future we intent to use other measures of sentence similarity. We also
plan to apply our algorithm to a wide spectrum of text segmentation tasks. We
are interested in segmentation of non artificial realife texts, texts having a
diverse distribution of segment length, long texts, and change-of-topic
detection in newsfeeds.

\section*{Acknowledgements}

The author would like to thank Georgios Orphanos for providing the POS Tagger
used for the preprocessing of the Greek texts and professors Athanassios
Kehagias and Vassilios Petridis for their undivided guidance and support.

\newpage
\begin{figure}
\begin{center}
{\includegraphics[width=4.16in,height=4.16in]{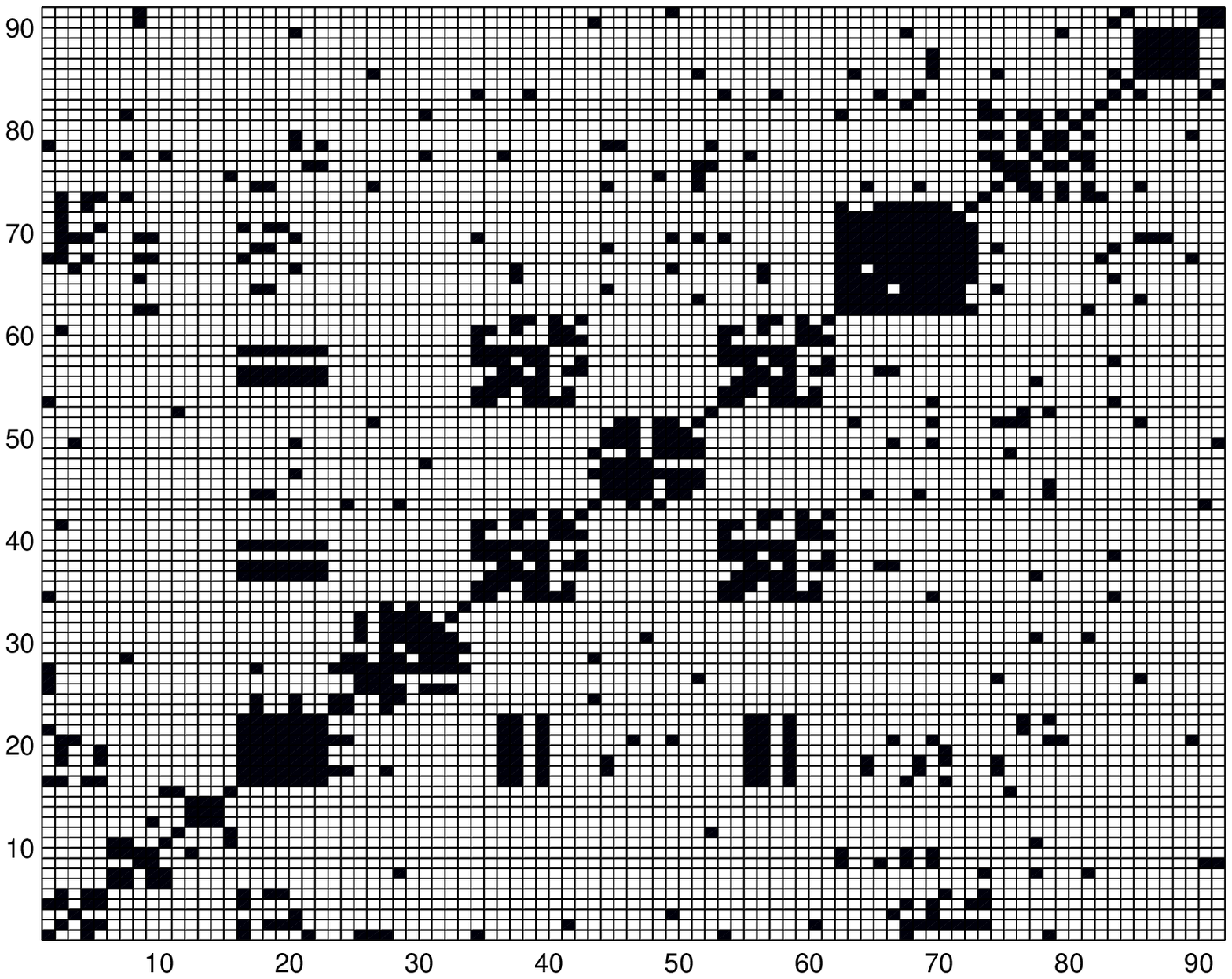} }%
\end{center}
\end{figure}
\textbf{Figure 1:} The similarity matrix D corresponding to a text
from the dataset '9-11' of Set 5. This text contains 91 sentences, hence D is
a 91 x 91 matrix. A black dot at position (m,n) indicates that the m-th and
n-th sentence have at least one word in common.

\begin{center}
\begin{tabular}{cc}
{\includegraphics[width=2.75in,height=3.00in]{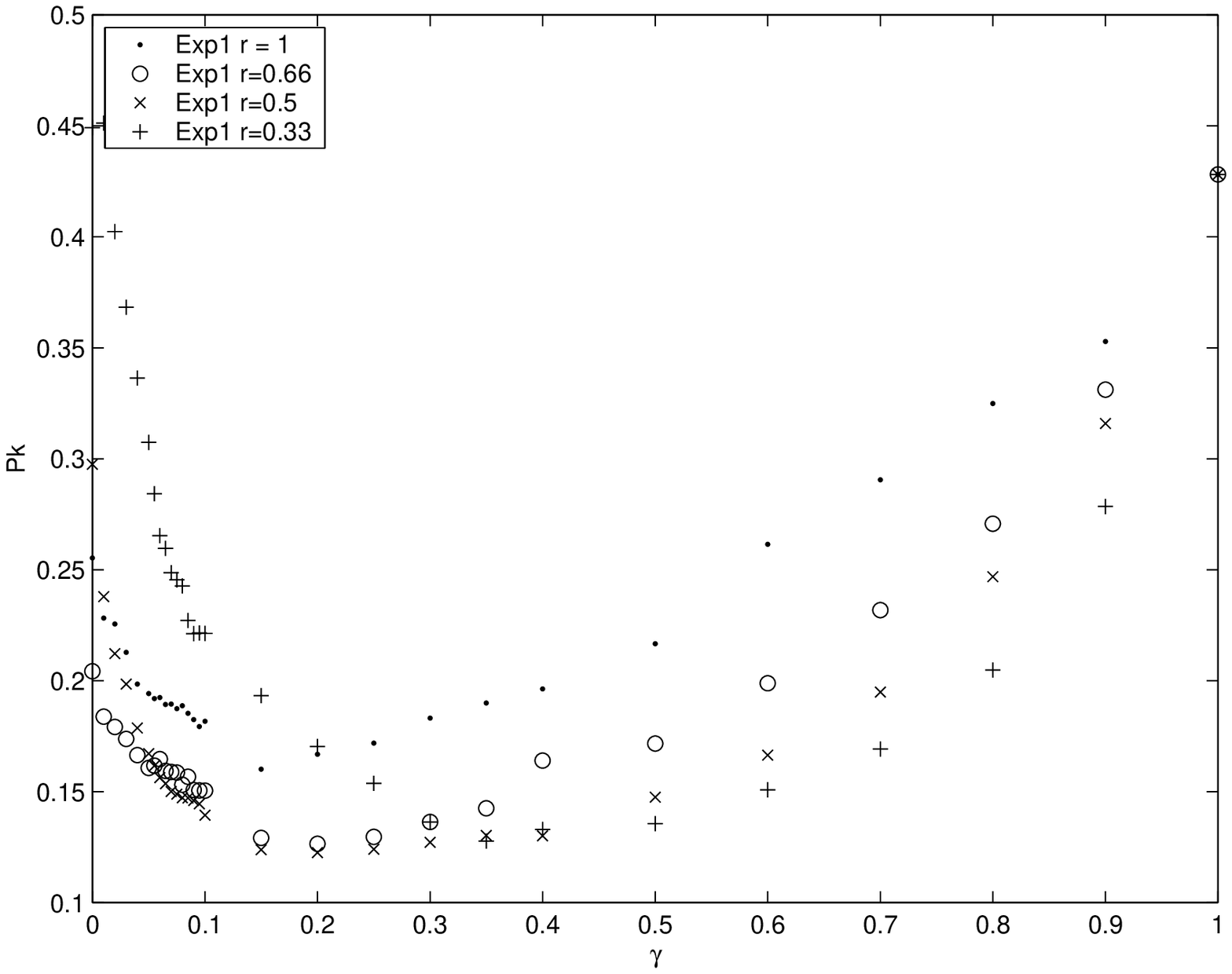} }  &  {\includegraphics[width=2.75in,height=3.00in]{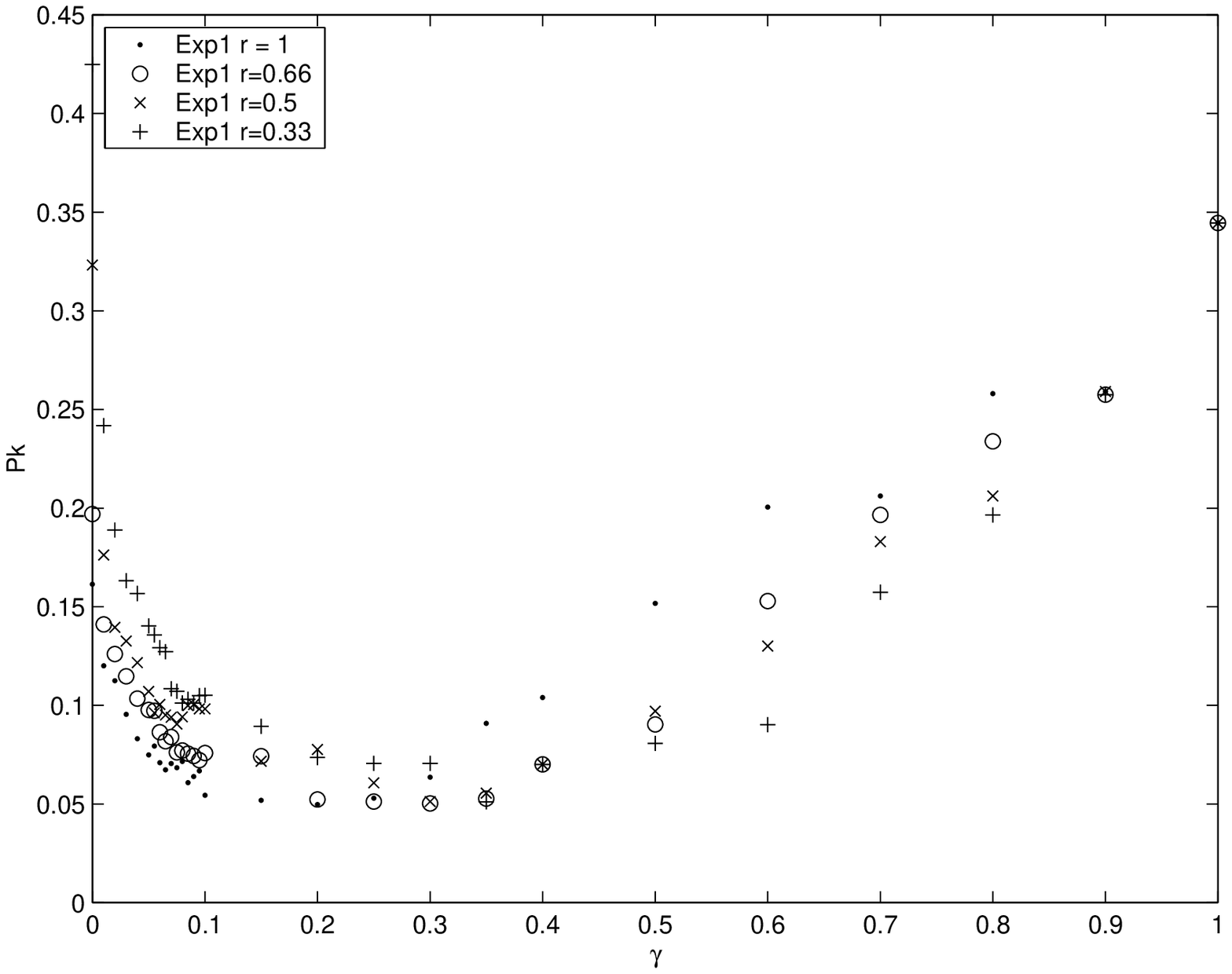} } \\ 
\textbf{Figure 2:}$P_{k}$ for '3-11' of Set5               &  \textbf{Figure 3:}$P_{k}$ for '3-5' of Set5                \\ 
 & \\
{\includegraphics[width=2.75in,height=3.00in]{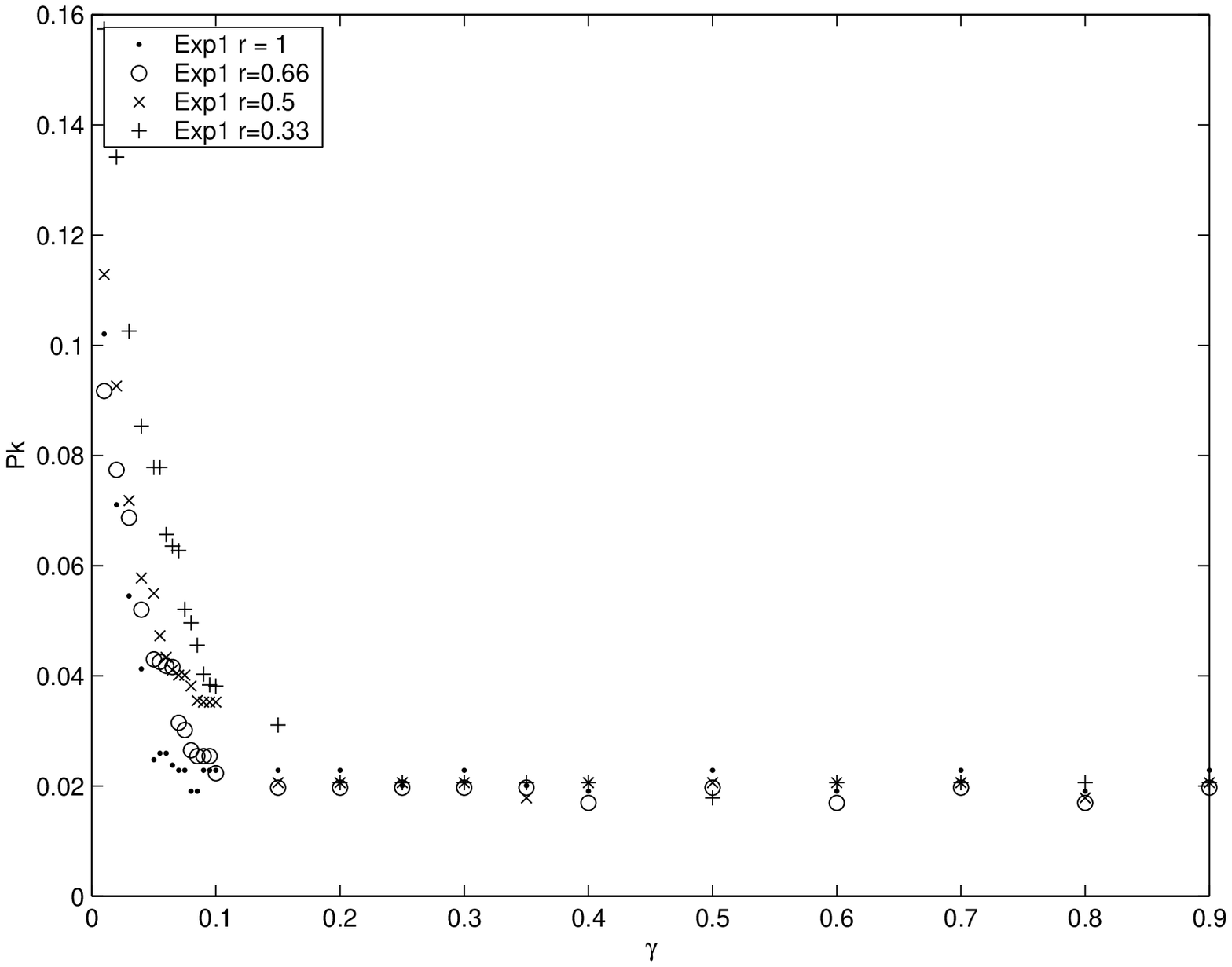} }  &  {\includegraphics[width=2.75in,height=3.00in]{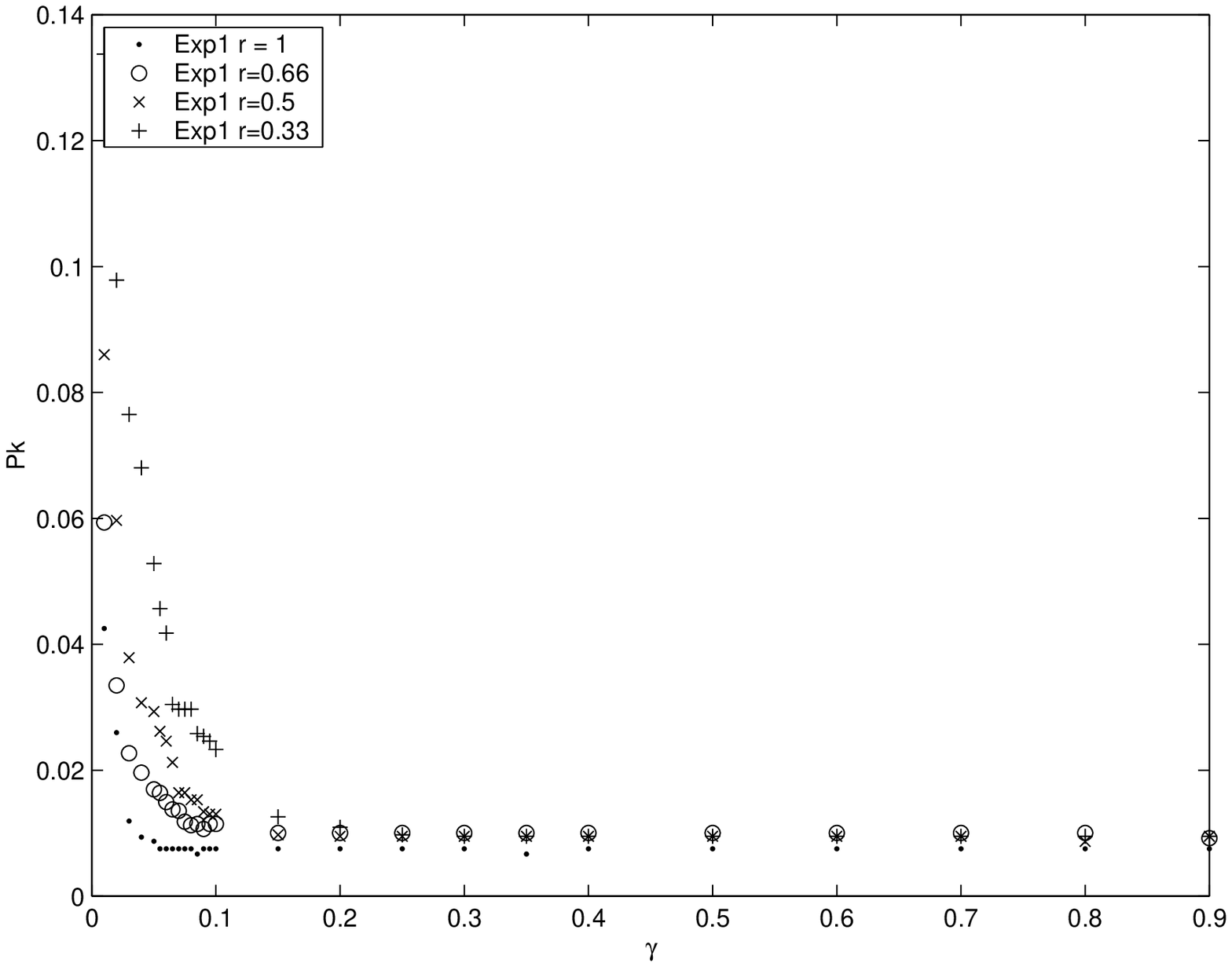} } \\ 
\quad \textbf{Figure 4:}$P_{k}$ for '6-8' of Set5          &  \quad \textbf{Figure 5:}$P_{k}$ for '9-11' of Set5
\end{tabular}
\end{center}


\begin{thebibliography}{99}
%{Orphanos \& Christodoulakis(1999)}                                                                %
\bibitem{Beeferman:97}Beeferman, D., Berger, A.\&
Lafferty, J. (1997). Text segmentation using exponential models. \emph{In
Proceedings of the 2nd Conference on Empirical Methods in Natural Language
Processing}, pp. 35-46.

\bibitem{Blei:01}Blei, D.M. \& Moreno, P.J. (2001).
Topic segmentation with an aspect hidden Markov model. Tech.Rep.\ CRL 2001-07,
COMPAQ Cambridge Research Lab.

\bibitem{Choi:00}Choi, F.Y.Y. (2000). Advances in domain
independent linear text segmentation. \emph{In Proceedings of the 1st Meeting
of the North American Chapter of the Association for Computational
Linguistics}, pp. 26-33.

\bibitem{Choi:01}Choi, F.Y.Y., Wiemer-Hastings, P.\&
Moore, J. (2001). Latent semantic analysis for text segmentation. \emph{In
Proceedings of the 6th Conference on Empirical Methods in Natural Language
Processing}, pp. 109-117.

\bibitem{Francis:82}Francis, W.N. \& Kucera, H.
(1982). \emph{Frequency Analysis of English Usage: Lexicon and Grammar}.
Houghton Mifflin.

\bibitem{Halliday:76}Halliday, M. \& Hasan, R.
(1976). \emph{Cohesion in English}. Longman.

\bibitem{Hearst:93}Hearst, M. A. \& Plaunt,C. (1993).
Subtopic structuring for full-length document access. \emph{In Proceedings of
the 16th Annual International of Association of Computer Machinery - Special
Interest Group on Information Retrieval (ACM / SIGIR) Conference on Research
and Development in Information Retrieval}, pp. 59-68.

\bibitem{Hearst:94}Hearst, M. A. (1994). Multi-paragraph
segmentation of expository texts. \emph{In Proceedings of the 32nd Annual
Meeting of the Association for Computational Linguistic}, pp. 9-16.

\bibitem{Heinonen:98}Heinonen, O. (1998). Optimal
Multi-Paragraph Text Segmentation by Dynamic Programming. \emph{In Proceedings
of 17th International Conference on Computational Linguistics}, pp. 1484-1486.

\bibitem{Hirschberg:93}Hirschberg, J. \& Litman,
D. (1993). Empirical studies on the disambiguation and cue phrases.
\emph{Computational Linguistics},vol.19, pp. 501-530.

\bibitem{Kan:98}Kan, M., Klavans, J.L. \& McKeown, K. R.
(1998). Linear segmentation and segment significance. \emph{In Proceedings of
the 6th International Workshop of Very Large Corpora}, pp. 197-205.

\bibitem{Kehagias:02}Kehagias, A., Petridis, V.,
Kaburlasos, V.G. \& Fragkou, P. (2002). \emph{\ A Comparison of Word- and
Sense-based Text Categorization Using Several Classification Algorithms}.
Journal of Intelligent Information Systems, Kluwer Academic Publishers vol.
21(3), pp. 227-247.

\bibitem{Kehagias:03}Kehagias, A , Fragkou, P. \&
Petridis, V. (2003). Linear Text Segmentation using a Dynamic Programming
Algorithm. \emph{In Proceedings of the European Association of Computational
Linguistics}, Budapest, Hungary, pp. 171-178.

\bibitem{Kozima:93a}Kozima, H. (1993). Text Segmentation based
on similarity between words. \emph{In Proceedings of the 31st Annual Meeting
of the Association for Computational Linguistics}, pp. 286-288.

\bibitem{Kozima:93b}Kozima, H \& Furugori, T.
(1993). Similarity between words computed by spreading activation on an
English dictionary. \emph{In Proceedings of 6th Conference of the European
Chapter of the Association for Computational Linguistics}, pp. 232-239.

\bibitem{Morris:91}Morris, J. \& Hirst, G. (1991).
\emph{Lexical cohesion computed by thesaural relations as an indicator of the
structure of text}. Computational Linguistics, vol.17, pp. 21-42.

\bibitem{Orphanos:99a}Orphanos, G. \&
Christodoulakis, D. (1999). Part-of-speech Disambiguation and Unknown Word
Guessing with Decision Trees, \emph{In Proceedings of EACL'99}, Bergen, Norway.

\bibitem{Orphanos:99c}Orphanos, G. \& Tsalidis,
C. (1999). Combining handcrafted and corpus-acquired Lexical Knowledge into a
Morphosyntactic Tagger. \emph{In Proceedings of the 2nd Research Colloquium
for Computational Linguistics in United Kingdom (CLUK)}, Essex, UK.

\bibitem{Passoneau:93}Passoneau, R. \& Litman,
D.J. (1993). Intention - based segmentation: Human reliability and correlation
with linguistic cues. \emph{In Proceedings of the 31st Meeting of the
Association for Computational Liguistics}, pp. 148-155.

\bibitem{Petridis:01}Petridis, V., Kaburlazos, V.,
Fragkou, P. \& Kehagias, A. (2001). Text Classification using the $\sigma
$-FLNMAP Neural Network. \emph{In Proceedings of the IJCNN'01 on Neural
Networks}.

\bibitem{Ponte:97}Ponte, J. M. \& Croft, W. B. (1997).
Text segmentation by topic. \emph{In Proceedings of the 1st European
Conference on Research and Advanced Technology for Digital Libraries}, pp. 120-129.

\bibitem{Reynar:98}Reynar, J.C. (1998). Topic Segmentation:
Algorithms and Applications. Ph.D. Thesis, Dept. of Computer Science, Univ. of Pennsylvania.

\bibitem{Reynar:99}Reynar, J.C. (1999). Statistical models for
topic segmentation. \emph{In Proceedings of the 37th Annual Meeting of the
Association for Computational Liguistics}, pp. 357-364.

\bibitem{Roget:77}Roget, P.M. (1977). \emph{Roget's
International Thesaurus}. Harper and Row, 4th edition.

\bibitem{Stamatatos:01}Stamatatos, E., Fakotakis,
N., \& Kokkinakis, G. (2001).\emph{Computer-Based Authorship Attribution
Without Lexical Measures}. Computer and the Humanities vol35, Kluwer Academic
Publishers, pp. 193-214.

\bibitem{Utiyama:01}Utiyama, M. \& Isahara, H.
(2001). A statistical model for domain - independent text segmentation.
\emph{In Proceedings of the 9th Conference of the European Chapter of the
Association for Computational Linguistics}, pp. 491-498.

\bibitem{Xu:96}Xu, J. \& Croft, W.B. (1996). Query
expansion using local and global document analysis. \emph{In Proceedings of
the 19th Annual International of Association of Computer Machinery - Special
Interest Group on Information Retrieval (ACM / SIGIR) Conference on Research
and Development in Information Retrieval}, pp. 4-11.

\bibitem{Yaari:97}Yaari, Y. (1997). Segmentation of expository
texts by hierarchical agglomerative clustering. \emph{In Proceedings of the
Conference on Recent Advances in Natural Language Processing}, pp. 59-65.

\bibitem{Yamron:99}Yamron, J. I., Carp, I.,Gillick,L.
Lowe, S. \& van Mulbregt P. (1999). Topic tracking in a news stream. \emph{In
Proceedings of DARPA Broadcast News Workshop}, pp. 133-136.

\bibitem{Youmans:91}Youmans, G. (1991). \emph{A new tool for
discourse analysis: The vocabulary management profile}. Language, vol. 67, pp. 763-789.
\end{thebibliography}
\end{document}